\documentclass{article}
\usepackage{ijcai20}

\usepackage{times}
\usepackage{soul}
\usepackage{url}
\usepackage[hidelinks]{hyperref}
\usepackage[utf8]{inputenc}
\usepackage[small]{caption}
\usepackage{graphicx}
\usepackage{amsmath}
\usepackage{amsthm}
\usepackage{booktabs}
\usepackage{algorithm}
\usepackage{algorithmic}
\usepackage{CJK}
\usepackage{arydshln}
\usepackage{multirow}
\usepackage{booktabs}
\usepackage{amsmath} 
\usepackage{amsfonts}
\usepackage{graphicx} 
\usepackage{subfigure}
\usepackage{amstext}
\usepackage{tikz}
\usepackage{pgfplots}

\title{Modeling Topical Relevance for Multi-Turn Dialogue Generation}

\author{Hainan Zhang$^{2}$\footnote{This work was done when the first author was a Ph.D student at ICT, CAS.} \and Yanyan Lan$^1$$^4$\footnote{Corresponding Author.} \and  Liang Pang$^1$$^4$ \and  Hongshen Chen$^2$ \and  Zhuoye Ding$^2$ \And  Dawei Yin$^4$ \\ 
\affiliations
$^1$CAS Key Lab of Network Data Science and Technology, Institute of Computing Technology, CAS \\
$^2$JD.com, Beijing, China \\$^3$Baidu.com, Beijing, China\\
$^4$University of Chinese Academy of Sciences
\emails
\{zhanghainan6,chenhongshen,dingzhuoye\}@jd.com,\\\{lanyanyan,pangliang\}@ict.ac.cn,yindawei@acm.org
}

\begin{document}

\maketitle

\begin{CJK*}{UTF8}{gbsn}
\begin{abstract}
 %\However, most existing generation models did not consider the topic information, but using all the contexts indiscriminately or attention-based contexts, leading to the inappropriate responses.
%which poses great challenges for designing the current dialogue generation models.  Therefore, an ideal generation model should be able to perceive the topic drift, detect the relevant contexts and generate appropriate responses accordingly. However, 

Topic drift is a common phenomenon in multi-turn dialogue. Therefore, an ideal dialogue generation models should be able to capture the topic information of each context, detect the relevant context, and produce appropriate responses accordingly. However, existing models usually use word or sentence level similarities to detect the relevant contexts, which fail to well capture the topical level relevance. In this paper, we propose a new model, named STAR-BTM, to tackle this problem. Firstly, the Biterm Topic Model is pre-trained on the whole training dataset. Then, the topic level attention weights are computed based on the topic representation of each context. Finally, the attention weights and the topic distribution are utilized in the decoding process to generate the corresponding responses. Experimental results on both Chinese customer services data and English Ubuntu dialogue data show that STAR-BTM significantly outperforms several state-of-the-art methods, in terms of both metric-based and human evaluations. 

\end{abstract}

\section{Introduction}
Multi-turn dialogue generation is widely used in many natural language processing (NLP) applications, such as customer services, mobile assistant and chatbots. Given a conversation history containing several contexts, a dialogue generation model is required to automatically output an appropriate response. Therefore, how to fully understand and utilize these contexts is important for designing a good multi-turn dialogue generation model. 

Different from single-turn dialogue generation, people usually model the multi-turn dialogue generation in a hierarchical way. A typical example is the Hierarchical Recurrent Encoder-Decoder (HRED) model ~\cite{SERBAN:HRED1,SORDONI:HRED2}. In the encoding phase, a recurrent neural network (RNN) based encoder is first used to encode each context as a sentence-level vector, and then a hierarchical RNN is utilized to encode these context vectors to a history representation. In the decoding process, another RNN decoder is conducted to generate the response based on the history representation. The parameters of both encoder and decoder are learned by maximizing the averaged likelihood of the training data. However, the desired response is usually only dependent on some relevant contexts, instead of all the contexts. Recently, some works have been proposed to model the relevant contexts by using some similarity measures. For example, Tian et al.~\shortcite{YAN:HARD} calculates the cosine similarity of the sentence embedding between the current context and the history contexts as the attention weights, Xing et al.~\shortcite{XING:SOFT} introduces the word and sentence level attention mechanisms to HRED, and Zhang et al.~\shortcite{ZHANG:RECOSA} utilizes the sentences level self-attention mechanism to detect the relevant contexts. However, these similarities are defined on either word or sentence level, which cannot well tackle the topic drift problem in multi-turn dialogue generation. 

\begin{table}
\centering
\scriptsize
\newcommand{\tabincell}[2]{\begin{tabular}{@{}#1@{}}#2\end{tabular}}
\begin{tabular}{ll} 
%\toprule
%\multicolumn{2}{c}{The first example}\\
\toprule
context1 & \tabincell{l}{你好，在吗？(Hello)}\\ 
context2 & \tabincell{l}{有什么问题我可以帮您呢? (What can I do for you?)}\\ 
context3 & \tabincell{l}{商品降价了，我要申请\textcolor{blue}{保价}\\(The product price has dropped. I want a \textcolor{blue}{low-price}.)}\\ 
context4 & \tabincell{l}{好的，这边帮您\textcolor{blue}{申请}，商品已经收到了吧？\\ (Ok, I will \textcolor{blue}{apply} for you. Have you received the product?)}\\ 
context5 & \tabincell{l}{东西收到了\textcolor{red}{发票}不在一起吗？\\  (I have received the product without the \textcolor{red}{invoice} together.)}\\ 
context6 & \tabincell{l}{开具\textcolor{red}{电子发票}不会随货寄出 \\  (The \textcolor{red}{electronic invoice} will not be shipped with the goods.)}\\ 
current context & \tabincell{l}{是发我\textcolor{red}{邮箱}吗？(Is it sent to my \textcolor{red}{email}？)}\\ \hline
response & \tabincell{l}{是的，请您提供邮箱地址，电子发票24小时寄出。\\ (Yes, please provide your email address，\\we will send the electronic invoices in 24 hours.)}\\ 
%\toprule
%\multicolumn{2}{c}{The second example}\\
%\toprule
%context1 & \tabincell{l}{Can someone help me ? My newli installed %in ubuntu can\\ not connect to my router}\\ 
%context2 & \tabincell{l}{Do you get an ip address?}\\ 
%context3 & \tabincell{l}{Yeah}\\ 
%context4 & \tabincell{l}{Then you should be connect : ) What your %problem ?}\\ 
%context5 & \tabincell{l}{Yup , it is 192.168.2.132 and I can %\textcolor{red}{ping} but not the router}\\ 
%context6 & \tabincell{l}{Your router is not the %\textcolor{red}{dhcp server} ?}\\ 
%current context & \tabincell{l}{If I try to %\textcolor{red}{restart} network through terminal , I lose the ip %\\address , but if I \textcolor{red}{restart} the whole computer %, I get one}\\ \hline
%response & \tabincell{l}{Well if the ip adress is assigned by the %\textcolor{red}{dhcp server} , \\you should be able tope ...}\\ 
\bottomrule
\end{tabular}
\caption{The example from the customer services dataset. The word color indicates the relevant topic word in the contexts and response, showing the topic-drift phenomenon.}\label{tb:examples}
%\caption{The example from the customer services dataset and ubuntu dataset. The word color indicates the relevant topic word in the contexts and response, showing the topic-drift phenomenon.} \label{tb:examples}
\end{table}
Here we give an example conversation, as shown in Table~\ref{tb:examples}. The contexts are of three different topics. The (context1,context2) pair talks about \lq{}greeting\rq{}, the (context3,context4) pair talks about \lq{}low-price\rq{}, and the (context5,...,response) pair talks about \lq{}invoice\rq{}. In this case, using all the contexts indiscriminately will obviously introduce many noises to the decoding process, which will hurt the performance of the multi-turn dialogue generation model. If we use word level similarities to locate the relevant contexts, the current context and context4 in the example will be associated because \lq{}send\rq{} and \lq{}receive\rq{} are highly similar words, which is clearly wrong. If we use sentence level similarities to locate the relevant contexts, it may still involve the false relevant context4 into consideration.

We argue that context relevance should be computed at the topic level, to better tackle the topic drift problem in multi-turn dialogue generation. From both linguistic and cognitive perspective, topic is the high level cluster of knowledge, which can describe the relationship of sentences in the context, and has an important role in human dialogue for directing focus of attention. In this paper, we propose a new model, namely STAR-BTM, to model the Short-text Topic-level Attention Relevance with Biterm Topic Model (BTM)~\cite{yan2013biterm}. Specifically, we first pre-train the BTM model on the whole training data, which split every customer-server pair in the context as a short document. Then, we use the BTM to get each sentence topic distribution and calculate the topic distribution similarity between the current context and each history context as the relevance attention. Finally, we utilize the relevance attention and the topic distribution to conduct the decoding process. The BTM model and the text generation model are jointly learned to improve their performances in this process.

In our experiments, we use two public datasets to evaluate our proposed models, i.e.,~Chinese customer services and English Ubuntu dialogue corpus. The experimental results show that STAR-BTM generates more informative and suitable responses than traditional HRED models and its attention variants, in terms of both metric-based evaluation and human evaluation. Besides, we have shown the relevant attention words, indicating that STAR-BTM obtains coherent results with human\rq{}s understanding.

\section{Related Work}
Recently, multi-turn dialogue generation has gained more attention in both research community and industry, compared with the single-turn dialogue generation~\cite{LI:GAN,MOU:BACKFORWARD,ZHANG:COHER,ZHANG:CVAR}. One of the reasons is that it is closely related to the real application, such as chatbot and customer service. More importantly, multi-turn dialogue generation needs to consider more information and constraints~\cite{CHEN:MVHRED,ZHANG:2018,ZHANG:RECOSA,WU:IR,ZHOU:IR}, which brings more challenges for the researchers in this area.
To better model the historical information, Serban et al. ~\cite{SERBAN:HRED1} propose the HRED model, which uses a hierarchical encoder-decoder framework to model all the contexts information. With the widespread use of HRED, more and more variant models have been proposed. For example, Serban et al.~\cite{SERBAN:VHERD,SERBAN:MRRNN} propose Variable HRED (VHRED) and MrRNN which utilize the latent variables as intermediate states to generate diverse responses. 
%VHRED uses a hidden variable as a semantic representation. By adding a Gaussian distribution constraint to a hidden variable, more robust text generation can be obtained. Based on HRED, the model adds a hidden variable constraint and generates a hidden variable. Growth distance dependent response.

However, it is unreasonable to use all the contexts indiscriminately for the multi-turn dialogue generation task, since the responses are usually only associated with a portion of the previous contexts. Therefore, some researchers try to use the similarity measure to define the relevance of the context. Tian et al.~\cite{YAN:HARD} propose a weighted sequence (WSeq) attention model for HRED, which uses the cosine similarity as the attention weight to measure the correlation of the contexts. 
%Specifically, they calculate the cosine similarity using the current context representation and each history context representation, and use the normalized similarity as the attention weight. We can see that 
But this model only uses the unsupervised sentence level representation, which fails to capture some detailed semantic information. Recently, Xing et al.~\cite{XING:SOFT} introduced the traditional attention mechanism~\cite{BAHDANAU:ATTENTION} into HRED, named hierarchical recurrent attention network (HRAN). In this model, the weight of attention is calculated based on the current state, the sentence level representation and the word level representation. However, the word level attention may introduce some noisy relevant contexts. Shen et al.~\cite{CHEN:MVHRED} propose to introduce the memory network into the VHRED model, so that the model can remember the context information. Theoretically, it can retrieve some relevant information from the memory in the decoding phase, however, it is not clearly whether and how the system accurately extracts the relevant contexts. Zhang et al.~\cite{ZHANG:RECOSA} proposed to use the sentence level self-attention to model the long distance dependency of contexts, to detect the relevant context for the multi-turn dialogue generation. Though it has the ability to tackle the position bias problem, the sentence level self-attention is still limited in capturing the topic level relevant contexts. 

The motivation of this paper is to detect the topic level attention relevance for multi-turn dialogue generation. It is a more proper way to deal with the topic draft problem, as compared with the traditional word or sentence level methods. Some previous works\cite{XING:TOPIC,XING:SOFT} have been proposed to use topic model in dialogue generation. They maily use the topic model to provide some topic related words for generation, while our work focuses on detecting the topic level relevant contexts.
%Different from previous studies, our proposed model can focus on the topic-level relevant contexts, by using the short-text topic model.

\section{STAR-BTM}
\begin{figure}[!t]
\centering
\includegraphics[width=0.5\textwidth]{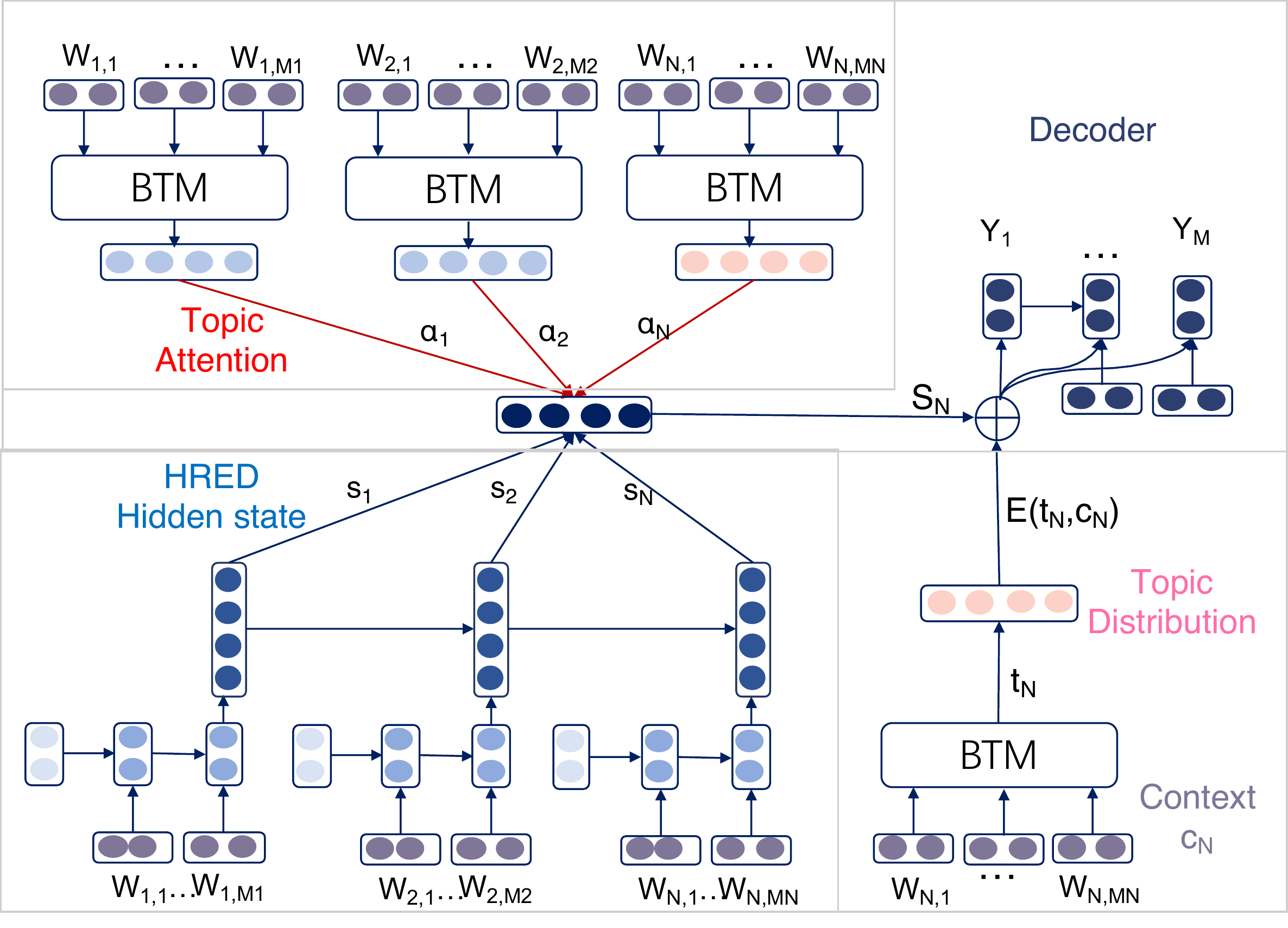}
\caption{ \label{fig:architecture} The architecture of STAR-BTM. }
\end{figure}

In this section, we will describe our Short-text Topic Attention Relevance with Biterm Topic Model (STAR-BTM) in detail, with the architecture shown in Figure~\ref{fig:architecture}. STAR-BTM consists of three modules, i.e.~, the pre-trained BTM model, topic level attention module and the joint learning decoder. 
Firstly, we pre-train the BTM model on the whole training data, to obtain the topic word distribution of each context. Secondly, the topic level attention is calculated as the similarity between the topic distributions of the current context and each history context. After that, the attention weights are multiplied with the hierarchical hidden state in HRED to obtain the history representation. Finally, the history representation and the topic distribution of the current context are concatenated to decode the response step by step.

From the architecture, we can see that STAR-BTM introduces the short text topic model into the HRED model, to incorporate the topic level relevant contexts to the decoding process. It is clear that the topic level distribution can provide more specific topic information than only using the word and sentence level representations. What is more, the topic model firstly \lq{}sees\rq{} the whole data globally by the pre-training techniques, and is then fine-tuned by the joint learning technique with the generation model. 

\subsection{Pre-train BTM Module}
We use the pre-trained BTM model on the whole training data to obtain the topic distribution. The pre-trained model on training data can be viewed as the background knowledge, which supplies additional information for the current dialogue session. Like human dialogue in reality, the background knowledge about potential topics will help to detect actual focus of attention model. 

BTM~\cite{yan2013biterm} is a widely used topic model especially designed for short text, which is briefly introduced as follows. For each co-occurrence biterm $b=(w_i,w_j)$ of word $w_i$ and $w_j$, the joint probability of $b$ is written as:
\begin{equation*}
P(b)=\sum_{t}P(t)P(w_i|t)P(w_j|t),
\end{equation*}
where $t$ stands for a topic.

To infer the topics in a document, BTM assumes that the topic
proportions of a document equals to the expectation of the
topic proportions of biterms generated from the document. Then we have,
\begin{equation}
P(t|d)=\sum_{b}P(t|b)P(b|d),\label{eq:topic}
\end{equation}
where $d$ is a document.

Both $P(t|b)$ and $P(b|d)$ can be calculated via Bayes’ formula as follows.
\begin{align*}
P(t|b)&=\frac{P(t)P(w_i|t)P(w_j|t)}{\sum_{t}P(t)P(w_i|t)P(w_j|t)}, \\
P(b|d)&=\frac{n_d(b)}{\sum_{b}n_d(b)},  
\end{align*}
where $n_d(b)$ is the frequency of the biterm $b$ in the document $d$. 
The parameters inference is based on the Gibbs Sampling. 

Now we introduce how we apply BTM in our work. Firstly, we split the whole training data $\mathcal{D}=\{(C,Y)=(c_1,\dots,c_N,Y)\}$ to context pairs, i.e.~$\mathcal{D}=\{(c_1,c_2),(c_3,c_4),\dots,(c_N,Y)\}$. In the training process, we treat each context pair as one document for BTM. This is reasonable because each pair can be viewed as a single-turn dialogue, and the input and output of a single-turn dialogue are usually of the same topic. After utilizing the Gibbs Sampling, we obtain the word distribution of each topic $P(w_i|t)$ and the topic distribution $P(t)$. In the inferring process, given each sentence $c_i$ in $\mathcal{D}$, the topic of $c_i$ is computed by $P(t_i)=\arg\max_{t}P(t|c_i)$ in Equation~\ref{eq:topic}.

The BTM model is more suitable for the dialogue generation task than the traditional topic models, such as Latent Dirichlet Allocation(LDA) model. That is because the dialogue has the characteristic of short text with omitted information, which makes LDA not reliable any more. BTM uses word co-occurrence as the core information to determine the topic. So it only depends on the semantic dominance of local co-occurrence information, breaks the document boundary, uses the information of the entire corpus instead of a single document to overcome the sparse problem in short text topic modeling. 

%The core idea is to directly model the global co-occurrence relationship. The advantage of this is that the word co-occurrence relationship contains context information, which is easier to judge the topic attribution of the word than the single word. More importantly, the word co-occurrence relationship has nothing to do with the document length. It learns the subject of short text by directly modeling the generation of biterms in the entire corpus. Here, biterm is an unordered word pair which co-occurs in a short context. BTM assumes that two words in a biterm share the same subject extracted from a mixed topic in the entire corpus. The topic here is also expressed as the word distribution of the traditional topic model. 

%Compared with the traditional topic model, the main differences and advantages of BTM are: 1) BTM explicitly simulates the co-occurrence mode, rather than implicit (through document modeling) to enhance the topic learning; 2) BTM uses the co-occurrence mode of the aggregate words in the corpus for topic discovery, avoiding the problem of document-level sparsity.

\subsection{Topic-level Attention Module}
We define the context data as $C = \{c_1,\dots,c_N \}$, and each sentence in $C$ is defined as $c_i = \{x^{(i)}_1,\dots,x^ {(i)}_M \}$. Given the sentence $c_i$ as input, the RNN model first maps the input sequence $c_i$ to the fixed dimension vector $h^i_M$ as follows:
\begin{equation*}
h^{(i)}_k=f(h^{(i)}_{k-1}, w^{(i)}_k),
\end{equation*}
where $w^{(i)}_k$ is the word vector of $x^{(i)}_k$, $h^{(i)}_k$ represents the hidden state vector of the RNN at time $k$, which combines $ w^{(i)}_k$ and $h^{(i)}_{k-1}$.
We obtained the state representation set of the contexts $\{h^{(1)}, \dots,h^{(N)} \}$.

Then we use a high-level RNN model to take the context state representation set $\{h^{(1)}, \dots,h^{(N)} \}$ as input, and obtain the high-level context representation vector $s_k$:
\begin{equation*}
s_k=f(s_{k-1},h^{(k)}),
\end{equation*}
where $h^{(k)}$ is the vector representation of the $k$-th sentence, and $s_k$ represents the state vector of the high-level RNN at time $k$, which combines $h^{(k)}$ and $s_{k-1}$.
We obtained the output of the high-level RNN at each step: $\{s_{1}, \dots, s_{N} \}$. 

Given the context data $C = \{c_1,\dots,c_N \}$, we obtained the topic for each sentence as $T = \{t_1,\dots,t_N \}$ through the pre-trained BTM model. We define attention weights as:
\begin{equation*}
\alpha_{i}=\frac{E(t_i,c_i) E(t_N,c_N) }{ |E(t_i,c_i)| \cdot | E(t_N,c_N)|},
\end{equation*}
where $E(t_i,c_i)$ is the sum of the word distribution for topic $t_i$ and the projected word distribution for context $c_i$, which is defined as the product of the word distribution for topic $t_i$ and the one-hot representation of context $c_i$.

%The word distribution for topic $t_i$ is $D_i= \{d_i^{(1)}, \dots, d_i^{(V)} \}$, and the mapping word distribution of $c_i$ is the product of $D_i$ and the one-hot representation of the context $c_i$. 

Finally, we obtain the softmax attention weights $\alpha_{i}'$ and the context vector $S_N$ as:
\begin{equation*}
\alpha_{i}'=\frac{\alpha_{i}}{\sum_{j=1}^{N} \alpha_{j}}, \;\;
S_N=\sum_{i=1}^{N}\alpha_{i}' \times s_{i}.
\end{equation*}

\subsection{Joint Learning Decoder}
We conduct another RNN as the decoder to generate the response $Y=\{y_1,\cdots, y_M\}$.
Given the context vector $S_N$, the topic distribution of the current context $D_N$ and the previously generated word ${y_1,\cdots,y_{i-1}}$, the decoder predicts the probability of the next word $y_i$ by converting the joint probability into a conditional probability through a chain rule in probability theory. We use the topic distribution of the current context $D_N$ in decoder for the reason that it could supply the topic information to generate more relevant response.

Given a set of training data $\mathcal{D}=\{(C,T,Y)\}$, STAR-BTM assumes that the data is conditionally independent, and samples from the probability $P_g$, and uses the following negative log likelihood as a minimized objective function:
\begin{equation*}
\mathcal{L}= -\sum_{(C,T,Y)\in\mathcal{D}} \log P_{g}(Y|C,T),
\end{equation*}
where $C$ is the context, $T$ is the topic distribution of $C$ and $Y$ is the real response.

\section{Experiment}
In this section, we conducted experiments on the Chinese customer service dataset and the English Ubuntu conversation dataset to verify the effectiveness of our proposed method.

\subsection {Experimental Settings}
We first introduce experimental settings, including datasets, baselines, parameter settings, and evaluation measures.

\subsubsection{Datasets}
We utilize two public multi-turn dialogue datasets in our experiments, which are widely used in the evaluation of multi-turn dialogue generation task. The Chinese customer service dataset, named JDC, consists of 515,686 history-response pairs published by the JD contest \footnote{\url{http://jddc.jd.com/}}. We randomly divided the corpus into training, validation and testing, each contains 500,000, 7843, and 7843 pairs, respectively.
The Ubuntu conversation dataset \footnote {\url{https://github.com/rkadlec/ubuntu-ranking-dataset-creator}} is extracted from the Ubuntu Q\&A forum, called Ubuntu~\cite{UBUNTU:2015}. 
%The original training data includes 7 million conversations from 2004 to Apr. 27, 2012. The validation set includes dialogue from Apr. 27, 2012 to Aug. 7, 2012, and the test set includes dialogue from Aug. 7, 2012 to Dec. 1, 2012. 
We utilize the official scripts for tokenizing, stemming and morphing, and remove the duplicates and sentence whose length is less than 5 or greater than 50. Finally, we obtain 3,980,000, 10,000, and 10,000 history-response pairs for training, validation and testing, respectively.

\subsubsection{Baseline Methods and Parameter Settings}
We used seven baseline methods for comparison, including the traditional Seq2Seq~\cite{SUTSKEVER:S2S}, HRED~\cite{SERBAN:HRED1}, VHRED~\cite{SERBAN:VHERD}, Weighted Sequence with Concat (WSeq)~\cite{YAN:HARD}, Hierarchical Recurrent Attention Network (HRAN)~\cite{XING:SOFT}, Hierarchical Hidden Variational Memory Network (HVMN)~\cite{CHEN:MVHRED} and Relevant Context with Self-Attention (ReCoSa)~\cite{ZHANG:RECOSA}. To fairly compare the topic-level attention model with self-attention model, we extend our STAR-BTM to the ReCoSa scenario, named ReCoSa-BTM, where the topic embedding is concatenated with the sentence representation.

For JDC, the Jieba tool is utilized for Chinese word segmentation, and its vocabulary size is set to 68,521. For Ubuntu, we set the vocabulary size to 15,000. To fairly compare our model with all baselines, the number of hidden nodes is all set to 512 and the batch size set to 32. The max length of sentence is set to 50 and the max number of dialogue turns is set to 15. The number of topics in BTM is set to 8. We use the Adam for gradient optimization in our experiments. The learning rate is set to 0.0001. We run all models on the Tesla K80 GPU with Tensorflow.\footnote{\url{https://github.com/zhanghainan/STAR-BTM}}

\subsubsection{Evaluation Measures}
We use both quantitative evaluation and human judgements in our experiments. Specifically, we use the traditional indicators, i.e., PPL and BLEU~\cite{XING:TOPIC} to evaluate the quality of generated responses~\cite{CHEN:MVHRED,YAN:HARD,XING:SOFT}. And we also use the {\em distinct} value~\cite{LI:MMI,LI:DRL} to evaluate the degree of diversity of generation responses. They are widely used in NLP and multi-turn dialogue generation tasks~\cite{CHEN:MVHRED,YAN:HARD,XING:SOFT,ZHANG:2018,ZHANG:COHER,ZHANG:CVAR}.

For human evaluation, given the 300 randomly sampled context and its generated responses from all the models, we invited three annotators (all CS majored students) to compare the STAR-BTM model with the baseline methods, e.g. win and loss, based on the coherence of the generated response with respect to the contexts. In particular, the win tag indicates that the response generated by STAR-BTM is more relevant than the baseline model. In order to compare the informativeness of the response generated by the models, we also require the annotators to label the informativeness of each model. If the response generated by STAR-BTM is more informative than the baseline, the annotator will label 1, otherwise label 0.

\subsection {Experimental Results}
Experimental results on two datasets are demonstrate below.
%Now we demonstrate our experimental results on the two public datasets.

\begin{table}[!ht]
\centering
\scriptsize
\begin{tabular}{lcccc}
\multicolumn{5}{c}{JDC Dataset}\\
\toprule
Model & PPL & BLEU & distinct-1 & distinct-2 \\ \hline
SEQ2SEQ & 20.287 & 11.458 & 1.069 & 3.587 \\
HRED & 21.264 & 12.087 & 1.101 & 3.809 \\
VHRED & 22.287 & 11.501 & 1.174 & 3.695 \\ \hline
WSeq & 21.824 & 12.529 & 1.042 & 3.917 \\
HRAN & 20.573 & 12.278 & \bf{1.313} & 5.753 \\
HVMN & 22.242 & 13.125 & 0.878 & 3.993 \\ \hline
STAR-BTM & \bf{20.267} & \bf{13.386} & 0.937 & \bf{5.816} \\ \hline
ReCoSa &  17.282 & 13.797 & 1.135 & 6.590 \\ 
ReCoSa-BTM & 18.432 & \bf{13.912} & 1.180 & \bf{6.739}  \\
\bottomrule
\multicolumn{5}{c}{Ubuntu Dataset}\\
\toprule
Model &PPL & BLEU & distinct-1 & distinct-2 \\ \hline
SEQ2SEQ & 104.899 & 0.4245 & 0.808 & 1.120 \\
HRED & 115.008 & 0.6051 & 1.045 & 2.724\\
VHRED & 186.793 & 0.5229 & 1.342 & 2.887\\ \hline
WSeq & 141.599 & 0.9074 & 1.024 & 2.878\\
HRAN & 110.278 & 0.6117 & 1.399 & 3.075\\
HVMN & 164.022 & 0.7549 & 1.607 & 3.245\\ \hline
STAR-BTM & \bf{104.893} & \bf{1.3303} & 1.601 & \bf{4.525} \\ \hline
ReCoSa & 96.057 & 1.6485 & 1.718 & 3.768  \\ 
ReCoSa-BTM & 96.124 & \bf{1.932} & \bf{1.723} & \bf{4.734} \\
\bottomrule
\end{tabular}
\caption{\label{tb:metricJDC} The metric-based evaluation results(\%). } % on JDC and Ubuntu
\end{table}

\begin{table}[!ht]
\centering
\scriptsize
\begin{tabular}{lcccc}
\multicolumn{5}{c}{JDC Dataset}\\
\toprule
\multirow{2}{1cm}{model} & \multicolumn{3}{c}{STAR-BTM vs.} & \multirow{2}{0.5cm}{kappa}\\ \cline{2-4}
 & win (\%) & loss (\%) & inform. (\%)\\ \hline
SEQ2SEQ & 55.32&2.12& 73.79&0.356\\
HRED & 48.93 & 6.38 & 70.87& 0.383\\
VHRED & 48.94&8.51& 69.98& 0.392\\ \hline
WSeq & 44.68& 8.5& 66.99& 0.378\\
HRAN & 34.04& 10.64& 60.19 & 0.401\\
HVMN & 27.66&12.77 & 61.02 & 0.379\\
ReCoSa & 25.34 & 20.71& 55.63 & 0.358  \\
\bottomrule
\multicolumn{5}{c}{Ubuntu Dataset}\\
\toprule
\multirow{2}{1cm}{model} & \multicolumn{3}{c}{STAR-BTM vs.} & \multirow{2}{0.5cm}{kappa}\\ \cline{2-4}
 & win (\%) & loss (\%) & inform. (\%)\\ \hline
SEQ2SEQ & 51.46& 3.88& 72.60& 0.398\\
HRED & 48.54& 6.80& 71.23 & 0.410\\
VHRED & 48.44& 6.76& 69.18 & 0.423\\ \hline
WSeq & 40.78 & 6.80& 67.80 & 0.415\\
HRAN & 32.04& 11.65& 61.16 & 0.422\\
HVMN & 25.24& 13.59& 60.27 & 0.414\\
ReCoSa & 20.14& 15.33 & 56.15& 0.409  \\
\bottomrule
\end{tabular}
\caption{\label{tb:humanEvaluation} The human evaluation on JDC and Ubuntu.}
\end{table}
\subsubsection{Metric-based Evaluation}
The metric-based evaluation results are shown in Table ~\ref{tb:metricJDC}. From the results, we can see that the models which detect the relevant contexts, such as WSeq, HRAN, HVMN and ReCoSa, are superior to the traditional HRED baseline models in terms of BLEU, PPL and {\em distinct}. This is mainly because these models further consider the attention of the relevant context information rather than all the contexts in the optimization process. HRAN introduces the traditional attention mechanism to learn the important context sentences. HVMN utilizes the memory network to remember the context information. ReCoSa uses the self-attention to detect the relevant contexts. But their effects are limited since they do not consider the topical level relevance. Our proposed STAR-BTM and ReCoSa-BTM have shown good results. Taking the BLEU value on the JDC dataset as an example, the BLEU value of the STAR-BTM and ReCoSa-BTM are 13.386 and 13.912, which are significantly better than that of HVMN and ReCoSa, i.e.,~13.125 and 13.797. The {\em distinct} value of our model is also higher than other baseline models, indicting that our model can generate more diverse responses. We also conducted a significance test. The results show that the improvement of our model is significant in both Chinese and English datasets with $\text {p-value} <0.01$. In summary, our STAR-BTM and ReCoSa-BTM model are able to generate higher quality and more diverse responses than the baselines.

\subsubsection{Human Evaluation}
The results of human evaluation are shown in Table ~\ref{tb:humanEvaluation}. The percentage of win, loss, and informativeness(inform.), as compared with the baseline models, are given to evaluate the quality and the informativeness of the generated responses by STAR-BTM. From the experimental results, the percentage of win is greater than the loss, indicating that our STAR-BTM model is significantly better than the baseline methods. Taking JDC as an example, STAR-BTM obtains a preference gain (i.e., the win ratio minus the loss ratio) of 36.18 \%, 23.4 \%, 14.89 \% and 4.63\%, respectively, as compared with WSeq, HRAN, HVMN and ReCoSa. In addition, the percentage of informativeness is more than 50 percent, as compared with WSeq, HRAN, HVMN and ReCoSa, i.e.,66.99\%， 60.19\%, 61.02\% and 55.63\%, respectively, showing that topic level information is effective for the multi-turn dialogue generation task and our STAR-BTM can generate interesting response with more information. 
The Kappa~\cite{KAPPA:1971} value demonstrates the consistency of different annotators.% We also conducted a significance test, and the improvement of our model is significant on both datasets, i.e., $\text {p-value} <0.01$.

%\begin{table*}[!t]
%\caption{\label{tb:ATTENTION} The attention analysis results(\%).}
%\centering
%\begin{tabular}{lccccccccc}
%\multicolumn{10}{c}{CSD Dataset}\\
%\toprule
%Model & P@1 & R@1 & F1@1& P@3 & R@3 &F1@3& P@5 & R@5 &F1@5 \\ \hline
%WSeq & \bf{35.20} & \bf{29.73} & \bf{16.12} & 24.27 & 51.49 & 16.50 & 21.61 & 71.76 & 16.61 \\
%HRAN & 22.88 & 15.56 & 9.26 & 24.13 & 46.22 & 15.85 & 22.78 & 66.22 & 16.95 \\ 
%ReCoSa & 27.56 & 20.35 & 11.71 & 26.20 & \bf{54.07} & 17.65 &23.92 & \bf{73.84} & \bf{18.07} \\ \hline
%STAR-BTM & \bf{35.20} & \bf{29.73} & \bf{16.12} & \bf{27.47} & 49.76 & \bf{17.70} & \bf{24.41} & 67.63 & 17.93 \\
%\bottomrule
%\end{tabular}
%\end{table*}

\begin{table*}[!t]
\centering
\tiny
\newcommand{\tabincell}[2]{\begin{tabular}{@{}#1@{}}#2\end{tabular}}
\begin{tabular}{lll}
\multicolumn{3}{c}{Example1}\\
\toprule
context1 & \tabincell{l}{ 我收到货，一个镜子碎了}&\tabincell{l}{ I received the goods, a mirror was broken.}\\
context2 & \tabincell{l}{ 可以{\bf 申请售后}}&\tabincell{l}{ You can {\bf apply for after-sale}}\\
context3 & \tabincell{l}{ 可以安排{\bf 补发}吗？}&\tabincell{l}{ Can you {\bf send a new one for a replacement}?}\\
response & \tabincell{l}{您已经{\bf 申请售后}了吗？}&\tabincell{l}{ Have you ever {\bf apply for the after-sale}?}\\ \hline
Seq2Seq & \tabincell{l}{有什么能帮到您的吗?}&\tabincell{l}{What can I do for you? }\\
HRED & \tabincell{l}{亲爱哒，请您稍等哦，小红人这就帮您去查}&\tabincell{l}{ Dear, please wait, I will help you check}\\
VHRED & \tabincell{l}{您好，为了更好的为您服务，请您提供下京东的订单号}&\tabincell{l}{ Hello, in order to better serve you, please provide the order number}\\ \hline
WSeq & \tabincell{l}{ 可以的}&\tabincell{l}{ Yes}\\
HRAN & \tabincell{l}{这个订单是吗}&\tabincell{l}{ Is this order? }\\
HVMN & \tabincell{l}{ ORDERID 这个是吗?} & \tabincell{l}{ ORDERID, this is yours? }\\ 
ReCoSa & \tabincell{l}{ 可以的，为您申请呢} & \tabincell{l}{ Yes, I will apply for you! }  \\ \hline
STAR-BTM & \tabincell{l}{ 以提交的{\bf 售后服务单}为准的 }&\tabincell{l}{ Based on the submitted {\bf after-sales service form}}\\
ReCoSa-BTM &\tabincell{l}{ 可以的，需要申请售后换货，选择缺件} & \tabincell{l}{ Yes, you need apply after-sales and select lack }\\
\bottomrule
\multicolumn{3}{c}{Example2}\\
\toprule
context1 & \multicolumn{2}{l}{I set up a dhcp server on my ubuntu machine . Client connect to it and get a dhcp address . However , those client cann't connect the network} \\
context2 &  \multicolumn{2}{l}{Have you set the default gateway up in the dhcpd config ? Same subnet on all machine ?} \\
context3 & \multicolumn{2}{l}{Same subnet . How do I configure a default gateway ?}\\
response & \multicolumn{2}{l}{Option router 192.168.129.10 in your dhcpd.conf on the dhcp server}\\ \hline
Seq2Seq & \multicolumn{2}{l}{I don't know} \\
HRED & \multicolumn{2}{l}{I don't know what I have to do with the same time}\\
VHRED & \multicolumn{2}{l}{I have a lot of UNK}\\ \hline
WSeq & \multicolumn{2}{l}{I don't know , but I don't know what I do that}\\
HRAN &  \multicolumn{2}{l}{You can use the ubuntu server} \\
HVMN & \multicolumn{2}{l}{Subnet will work}\\ 
ReCoSa & \multicolumn{2}{l}{ How about the dhcp server?}  \\ \hline
STAR-BTM & \multicolumn{2}{l}{ Try restart it}\\
ReCoSa-BTM & \multicolumn{2}{l}{Dhcp server with router} \\
\bottomrule
\end{tabular}
\caption{\label{tb:caseJDC} The generated responses from the STAR-BTM model on JDC dataset. }
\end{table*}

\begin{table*}[!t]
\centering
\tiny
\newcommand{\tabincell}[2]{\begin{tabular}{@{}#1@{}}#2\end{tabular}}
\begin{tabular}{l|llllllllll}
\toprule
Topic & \multicolumn{10}{c}{Topic top 10 words in JDC dataset.}\\ \hline
1&\tabincell{l}{订单\\order}&\tabincell{l}{配送\\delivery}&\tabincell{l}{请\\please}&\tabincell{l}{商品\\item}&\tabincell{l}{时间\\time}&\tabincell{l}{站点\\site}&\tabincell{l}{联系\\contact}&\tabincell{l}{电话\\phone}&\tabincell{l}{亲爱\\dear}&\tabincell{l}{地址\\address }\\ \hline
2&\tabincell{l}{ 发票\\invoice}&\tabincell{l}{地址\\address}&\tabincell{l}{订单\\order}&\tabincell{l}{修改\\modification}&\tabincell{l}{电子\\electronic}&\tabincell{l}{开具\\issue}&\tabincell{l}{需要\\need}&\tabincell{l}{电话\\phone}&\tabincell{l}{号\\number}&\tabincell{l}{姓名\\name}\\ \hline
3&\tabincell{l}{ 工作日\\work day}&\tabincell{l}{退款\\refunds}&\tabincell{l}{订单\\order}&\tabincell{l}{账\\accounts}&\tabincell{l}{取消\\cancellations}&\tabincell{l}{申请\\applications}&\tabincell{l}{支付\\payments}&\tabincell{l}{成功\\successes}&\tabincell{l}{商品\\goods}&\tabincell{l}{请\\please}\\ \hline
4&\tabincell{l}{ 申请\\apply}&\tabincell{l}{售后\\after-sale}&\tabincell{l}{点击\\click}&\tabincell{l}{端\\end}&\tabincell{l}{提交\\submit}&\tabincell{l}{客户服务\\customer service}&\tabincell{l}{审核\\review}&\tabincell{l}{链接\\link}&\tabincell{l}{返修\\rework}&\tabincell{l}{补发\\replacement}\\  \hline
5&\tabincell{l}{ 订单\\order}&\tabincell{l}{站点\\site}&\tabincell{l}{时间\\time}&\tabincell{l}{日期\\date}&\tabincell{l}{下单\\order}&\tabincell{l}{ORDERID}&\tabincell{l}{编号\\number}&\tabincell{l}{催促\\urging}&\tabincell{l}{信息\\information}&\tabincell{l}{订单号\\order number}\\ \hline
6&\tabincell{l}{ 商品\\products}&\tabincell{l}{金额\\money}&\tabincell{l}{保价\\low-price}&\tabincell{l}{姓名\\name}&\tabincell{l}{手机\\mobile}&\tabincell{l}{申请\\apply}&\tabincell{l}{快照\\snapshot}&\tabincell{l}{订单\\order}&\tabincell{l}{查询\\inquiries}&\tabincell{l}{请\\please}\\ \hline
7&\tabincell{l}{查询\\inquire}&\tabincell{l}{帮\\help}&\tabincell{l}{调货\\delivery}&\tabincell{l}{问题\\problem}&\tabincell{l}{处理\\deal}&\tabincell{l}{缺货\\out of stock}&\tabincell{l}{订单号\\order number}&\tabincell{l}{提供\\offer}&\tabincell{l}{采购\\purchase}&\tabincell{l}{ 请\\please}\\ \hline
8&\tabincell{l}{!}&\tabincell{l}{帮到\\help}&\tabincell{l}{谢谢\\thank}&\tabincell{l}{支持\\support}&\tabincell{l}{感谢您\\thank you}&\tabincell{l}{评价\\evaluation}&\tabincell{l}{客气\\kind}&\tabincell{l}{妹子\\I}&\tabincell{l}{请\\please}&\tabincell{l}{祝您\\wish you }\\
\toprule
Topic & \multicolumn{10}{c}{ Topic top10 words in Ubuntu dataset}\\ \hline
1&	import& each& not& old& noth& would& than& of& thinic& retri \\ 
2&	cover& ad\-hoc& version& each& retri& alt& benefit& would& ubuntu& apt\_preferec \\ 
3&	from& cover& alt& or& consid& ed& link& we& window& minut\\ 
4&	run& desktop& cover& kick& distribut& browser& old& show& laptop& ars\\ 
5&	each& show& instead& from& irc& over& saw& rpm& mockup& out\\ 
6&	not& libxt-dev& big& a& by& reason& aha& cover& interest& !\\ 
7&	896& on& system& cover& restart& not& urgent& violat& overst& ping\\ 
8&	kxb& but& charg& alway& polici& f& \_my\_& aha& ugh& zealous\\ 
\bottomrule
\end{tabular}
\caption{\label{tb:TOPIC} The top10 words for each topic from the BTM model on JDC dataset. }
\end{table*}

\subsubsection{Case Study}
To facilitate a better understanding of our model, we present some examples in Table ~\ref{tb:caseJDC}, and show the top 10 words of each topic in the Table~\ref{tb:TOPIC}.
From the results, we can see that why the topic level attention model performs better than the model only using the word and sentence level representation. Taking the example1 in Table ~\ref{tb:caseJDC} as an example, it easy to generate common responses by using only sentence level representation, such as \lq{}{\em What can I do for you?}\rq{} and \lq{}{\em Yes}\rq{}. However, our topic level attention model has the ability to generate more relevant and informative responses, such as \lq{}{\em Based on the submitted after-sales service form}\rq{} and \lq{}{\em Yes, you need apply after-sales and select lack}\rq{}. This is mainly because the topic level attention is able to associate some important information such as \lq{}补发(send a new one for a replacement)\rq{} and \lq{}售后(after-sales)\rq{} by topic modeling, which are usually hard to be captured by traditional word or sentence level similarities. These results indicate the advantage of modeling topic level relevance.

We also show the top 10 words of each topic from the BTM model on the two dataset, as shown in Table ~\ref{tb:TOPIC}. Take the JDC dataset as an example, from the results, we can see that BTM model distinguishes eight topics, i.e., \lq{}配送(shipping), 发票(invoice), 退款(refund), 售后(after-sale), 催单(reminder), 保价(low-price),缺货(out-of-stock) and 感谢(thanks)\rq{}. For each topic, the top 10 words represent the core information of the topic. Take the example1 in the Table ~\ref{tb:caseJDC} as an example, since the \lq{}补发(send a new one for a replacement)\rq{} and \lq{}售后(after-sales)\rq{} are the 15-th and second word in the same topic 4, respectively, the model can generate \lq{}submitted after-sales service form\rq{} based on the topic level attention. In the example2, the current context is about the \lq{}gateway\rq{} with topic \lq{}network\rq{}, so the topic distribution can supply some additional topic information, such as \lq{}restart\rq{}, \lq{}dhcp\rq{} and \lq{}router\rq{}. In a word, our STAR-BTM and ReCoSa-BTM model can supply the critical topic information to improve the informativeness of the generated response.

\section{Conclusion}
In this paper, we propose a new multi-turn dialogue generation model, namely STAR-BTM. The motivation comes from the fact that topic drift is a common phenomenon in multi-turn dialogue. The existing models usually use word or sentence level similarities to detect the relevant contexts, which ignore the topic level relevance. Our core idea is to utilize topic models to detect the relevant context information and generate a suitable response accordingly. Specifically, STAR-BTM first pre-trains a Biterm Topic Model on the whole training data, and then incorporate the topic level attention weights to the decoding process for generation.We conduct extensive experiments on both Chinese customer services dataset and English Ubuntu dialogue dataset. The experimental results show that our model significantly outperforms existing HRED models and its attention variants. Therefore, we obtain the conclusion that the topic-level information can be useful for improving the quality of multi-turn dialogue generation, by using proper topic model, such as BTM. 

In future work, we plan to further investigate the proposed STAR-BTM model. For example, some personal information can be introduced to supply more relevant information for personalized modeling. In addition, some knowledges like concerned entities can be considered in the relevant contexts to further improve the quality of generated response.

\section*{Acknowledgments}
This work was supported by Beijing Academy of Artificial Intelligence (BAAI) under Grants No. BAAI2019ZD0306, and BAAI2020ZJ0303, the National Natural Science Foundation of China (NSFC) under Grants No. 61722211, 61773362, 61872338, and 61906180, the Youth Innovation Promotion Association CAS under Grants No. 20144310, and 2016102, the National Key R\&D Program of China under Grants No. 2016QY02D0405, the Lenovo-CAS Joint Lab Youth Scientist Project, and the Foundation and Frontier Research Key Program of Chongqing Science and Technology Commission (No. cstc2017jcyjBX0059), and the Tencent AI Lab Rhino-Bird Focused Research Program (No. JR202033).

\end{CJK*}

\bibliographystyle{named}
\bibliography{ijcai20}

\end{document}